%% file: main.tex
\documentclass[10pt,twocolumn,letterpaper]{article}

\usepackage[pagenumbers]{cvpr} 

\input{preamble}

\definecolor{cvprblue}{rgb}{0.21,0.49,0.74}
\usepackage[pagebackref,breaklinks,colorlinks,allcolors=cvprblue]{hyperref}

\def\paperID{8} 
\def\confName{CVPR}
\def\confYear{2026}

\title{CylinderDepth: Cylindrical Spatial Attention for Multi-View Consistent Self-Supervised Surround Depth Estimation}

\author{
Samer Abualhanud\quad Christian Grannemann \quad Max Mehltretter\\
Leibniz University Hannover\\
{\tt\small \{abualhanud, grannemann, mehltretter\}@ipi.uni-hannover.de}
}

\begin{document}
\maketitle
\input{sec/0_abstract}    
\input{sec/1_intro}

\input{sec/2_related_work}
\input{sec/3_method}

\input{sec/4_experiments}

\input{sec/5_conclusions}
{
    \small
    \bibliographystyle{ieeenat_fullname}
    \bibliography{main}
}


\end{document}

%% file: preamble.tex
\usepackage{booktabs}
\usepackage{multirow}
\usepackage{array}
\usepackage{caption}

%% file: sec/0_abstract.tex
\begin{abstract}
Self-supervised surround-view depth estimation enables dense, low-cost 3D perception with a 360$^\circ$ field of view from multiple minimally overlapping images. Yet, most existing methods suffer from depth estimates that are inconsistent across overlapping images. To address this limitation, we propose a novel geometry-guided method for calibrated, time-synchronized multi-camera rigs that predicts dense metric depth. Our approach targets two main sources of inconsistency: the limited receptive field in border regions of single-image depth estimation, and the difficulty of correspondence matching. We mitigate these two issues by extending the receptive field across views and restricting cross-view attention to a small neighborhood. To this end, we establish the neighborhood relationships between images by mapping the image-specific feature positions onto a shared cylinder. Based on the cylindrical positions, we apply an explicit spatial attention mechanism, with non-learned weighting, that aggregates features across images according to their distances on the cylinder. The modulated features are then decoded into a depth map for each view. Evaluated on the DDAD and nuScenes datasets, our method improves both cross-view depth consistency and overall depth accuracy compared with state-of-the-art approaches. Code is available at \url{https://abualhanud.github.io/CylinderDepthPage/}.
\end{abstract}
                                                                                                                                                                                                                                                                  

%% file: sec/1_intro.tex
\section{Introduction}
\label{sec:intro}
Depth estimation is an important step in 3D reconstruction and thus a crucial prerequisite for 3D scene understanding, enabling, for example, localization, obstacle avoidance and motion planning in autonomous driving and robotics. Due to the density of observations, the availability of radiometric information, and the comparably low cost, cameras are commonly used for this task. Recent learning-based depth estimation methods, often based on fully-supervised training, produce accurate and dense predictions. However, this requires ground-truth labels, often obtained by additional sensors such as LiDAR, yet, these labels are usually sparse. In contrast, self-supervised methods enforce photometric consistency between a target image and a rendered target image, generated by sampling pixels from a source image using the estimated depth and known camera parameters.\par

\begin{figure}[t]
    \centering
    
    \begin{subfigure}{0.495\linewidth}
        \centering
        \includegraphics[width=\linewidth]{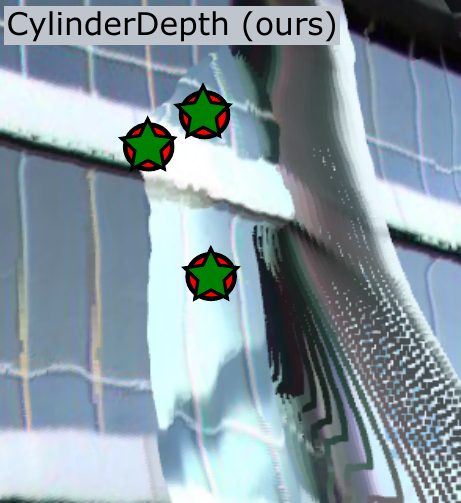}
    \end{subfigure}
    \begin{subfigure}{0.495\linewidth}
        \centering
        \includegraphics[width=\linewidth]{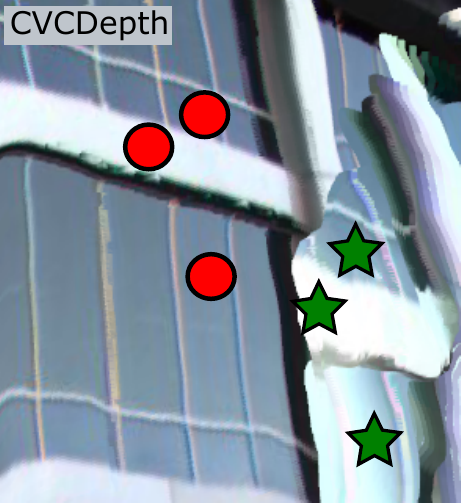}
    \end{subfigure}
    
    \caption{\normalsize Comparison of multi-view consistency between our method and CVCDepth~\cite{ding2024towards}. The stars and circles denote 3D reconstructions of the same 3D object points from two different images. While prior work struggles to achieve consistency in the reconstruction across images, our method clearly mitigates this limitation.}
    \label{fig:network}
\end{figure}

Surround camera setups, which consist of multiple calibrated cameras that are rigidly mounted to each other, provide a full 360$^\circ$ scene coverage and are widely used in autonomous driving \cite{caesar2020nuscenes, guizilini20203d}. In contrast to a single omnidirectional image, these setups allow for metric-scale depth estimation, given that the relative orientation parameters and the length of the baselines between the cameras are known. However, these setups typically provide only minimal spatial overlap. To address this, monocular temporal context is required to increase the effective overlap during training. However, processing each image independently can yield inconsistent depth estimates across cameras; a 3D object point that is visible in multiple images may get assigned different 3D coordinates per image, resulting in an inconsistent and misaligned reconstruction when combining the results obtained for the individual images. Most prior work enforces multi-view consistency only implicitly during training, e.g., by constraining motion to be consistent across cameras \cite{wei2023surrounddepth, kim2022self, ding2024towards}, adding loss functions that encourage consistency \cite{guizilini2022full, ding2024towards}, or using purely learned attention mechanisms \cite{wei2023surrounddepth, shi2023ega}. However, these approaches do not guarantee consistency at inference time, since the cameras’ geometric relationships are not considered.\par

To address this limitation, we propose a novel self-supervised depth estimation method for surround-view camera setups that enforces multi-view consistency by expanding the receptive field in border regions and constraining correspondence matching to a small neighborhood. Given the intrinsic and relative orientation parameters and an initial predicted depth, the 3D points reconstructed from all images are mapped onto a shared unit cylinder. This produces a unified representation across images in which pixels are indexed by cylindrical coordinates and where reconstructions of the same 3D point from multiple images are projected to the same 2D point on the cylinder. Thus, this projection establishes consistent neighborhood relations across images, aligning overlapping image regions. In contrast to approaches that exchange features between images without explicitly modeling their geometric relationship, typically using purely learned attention, we introduce a non-learned spatial attention weighting that weights pixel interactions based on the geodesic distances between their cylindrical coordinates. Thus, our main contributions are:
\begin{itemize}
    \item We propose a \textbf{spatial attention} mechanism for surround camera systems with \textbf{non-learned geometry-guided} weighting \item To enforce \textbf{multi-view consistency} during training and inference, we propose a mapping onto a shared \textbf{cylindrical representation}.
    \item We thoroughly evaluate our proposed method, focusing on \textbf{multi-view consistency}. In this context, we further present a novel \textbf{depth consistency metric}, closing a relevant gap in the literature.
\end{itemize}

%% file: sec/2_related_work.tex
\section{Related Work}
\label{sec:related}

\paragraph{Monocular Depth Estimation}

In monocular depth estimation, a dense, per‑pixel depth map is predicted from a single RGB image, which is an ill‑posed task. Learning semantic and geometric cues, supervised methods~\cite{ranftl2021vision, fu2018deep, liu2015learning, eigen2014depth, agarwal2023attention} rely on depth sensors for ground truth labels, which makes the sensor setup and its calibration more complex, while the obtained ground truth is often sparse. Self‑supervised approaches commonly optimize for photometric consistency across stereo image pairs~\cite{godard2017unsupervised, garg2016unsupervised}, image sequences~\cite{zhou2017unsupervised, godard2019digging, yin2018geonet, guizilini20203d, liu2024mono, watson2021temporal, mahjourian2018unsupervised, ruhkamp2021attention} or both~\cite{wimbauer2023behind, watson2019self}. However, these methods commonly focus on images with narrow fields of view, which are not sufficient to capture an entire scene. Addressing this limitation, another line of work employs omnidirectional images~\cite{vasiljevic2020neural, wang2018self}. However, all the aforementioned setups have no baselines, which does not allow for scale-aware self-supervised depth estimation.


\paragraph{Multi-View Depth Estimation}

Given multiple overlapping images, depth can be inferred through multi‑view stereo (MVS) reconstruction. Learning-based MVS methods can be grouped into two families: (i) methods based on the classical concept of photogrammetry, i.e., on the identification of image point correspondences and their triangulation to obtain 3D object points~\cite{gu2020cascade, im2019dpsnet, yao2018mvsnet, wang2022mvster, khot2019learning, yao2019recurrent}. (ii) pointmap regression methods, which directly predict 3D points, often together with the orientation parameters of the images~\cite{wang2024dust3r, leroy2024grounding, wang2025vggt}.
Typically, such MVS methods assume a 3D object point to be visible in two or more images, requiring sufficient overlap between the images either during training, inference or both.\par

\begin{figure*}[ht]
    \centering
    \includegraphics[width=1\textwidth]{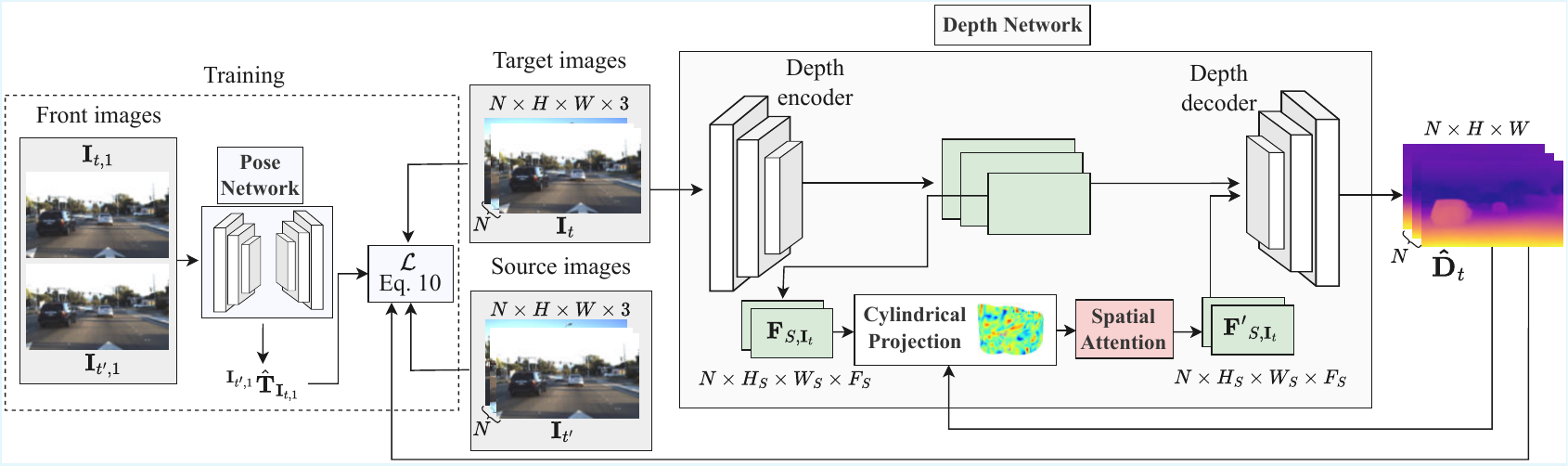}
    \caption{\normalsize Overview of the proposed network. The depth network takes the target images $\mathbf{I}{_{t}}$ as input.  The lowest-scale features $\mathbf{F}{_{S, \mathbf{I}_t}}$ from all target images are projected onto a cylinder, where attention is applied based on cylindrical distances. The pose network takes the source $\mathbf{I}{_{t', 1}}$ and target $\mathbf{I}{_{t,1}}$ front images as input to predict the relative metric pose between two frames.}
    \label{fig:netwossrk}
\end{figure*}

In contrast, multi-view surround camera setups provide a 360$^\circ$ field of view by combining multiple cameras, following the central projection model, with minimally overlapping image planes. 
Consequently, for the majority of pixels, depth needs to be estimated monoscopically. 
Recent work has studied this camera configuration for depth estimation, using both, images from a single~\cite{guizilini2022full, xu2025self, li2024m2depth, kim2022self, wei2023surrounddepth, ding2024towards, yang2024towards, shi2023ega} and from multiple  time steps~\cite{fei2024driv3r, zou2024m, schmied2023r3d3} during inference. The present work also focuses on this camera configuration, using images from a single time step during inference.
FSM~\cite{guizilini2022full} is among the earliest self-supervised methods for surround-view depth estimation. It leverages the spatio-temporal context for photometric supervision, exploits overlapping image regions to recover metric scale from a single time step, and introduces a loss to enforce consistency in the temporal pose prediction of the individual cameras. Subsequent work~\cite{wei2023surrounddepth, kim2022self} assumes a shared rigid motion of the camera rig and estimates the ego motion instead of the individual camera motion.
SurroundDepth~\cite{wei2023surrounddepth} proposes attention across images to enhance the consistency of the predicted depth maps. To obtain metric scale, a spatial photometric loss on overlapping images is combined with sparse pseudo-depth labels computed via SfM and filtered for outliers using epipolar geometry-based constraints.
In contrast, VFDepth~\cite{kim2022self} models the depth and pose as volumetric feature representations, i.e., operating in 3D instead of 2D space. However, 3D- and attention-based methods are computationally expensive and do not fully exploit the geometric relationships between images to enforce consistency at inference.

\paragraph{Attention-Based Depth Estimation}
Initially developed for natural language processing, attention is now widely used in vision-based tasks, including monocular~\cite{agarwal2023attention, ruhkamp2021attention, li2023depthformer, yun2022improving, guizilini2022multi, lee2021patch, johnston2020self, ranftl2021vision} and multi-view~\cite{luo2020attention, wang2022mvster, wei2023surrounddepth, shi2023ega} depth estimation.
Early progress was marked by DPT~\cite{ranftl2021vision}, which replaced conventional CNN backbones with Vision Transformers for dense prediction extending the receptive field. Attention can also be used to promote consistency in depth prediction. A work closely related to ours is~\cite{ruhkamp2021attention}, which employs spatial attention; however, it addresses multi-frame monocular depth estimation by aggregating features within each image based on pixel-wise 3D Euclidean distances, relying on estimated depth for the 3D projection, and further adds temporal attention to aggregate features across different time frames to enforce temporal consistency. Different from previous methods, we introduce a cross-view attention with non-learned spatial weighting that fuses features across images by making use of the geometric relations between the images.

%% file: sec/3_method.tex
\section{Methodology}
\label{sec:method}

Given a surround camera setup capturing $N$ time-synchronized images with spatial overlap and known intrinsic parameters and metric relative poses, i.e., known relative orientations and baselines in metric units between the cameras, we aim to estimate a depth map for every image. The depth network employed in our work follows an encoder–decoder architecture (see Fig.~\ref{fig:netwossrk}). In a first forward pass, input images $\mathbf{I}{_{t}} \in\mathbb{R}^{N\times H\times W\times 3}$ at time $t$, with $H$ and $W$ denoting the height and width of the images, respectively, are processed separately by a shared encoder to produce multi-scale feature maps $\mathbf{F}{_{s, \mathbf{I}_t}} \in\mathbb{R}^{N\times H_s\times W_s\times F_s}$, where $s \in \{1, \ldots, S\}$ is the scale, $H_s$ and $W_s$ are the height and width in $s$, respectively, and $F_s$ is the feature dimension. Passing these feature maps through the decoder, this first forward pass yields a preliminary depth prediction. In a second forward pass, we reuse the encoded feature maps and project their pixel positions onto a shared unit cylinder, based on the preliminary depth predictions and the known camera parameters. This enables feature aggregation via attention based on the pixels' geodesic distance on the cylinder to enforce consistent depth predictions across images (see Sec.~\ref{cylproj}). We apply the proposed spatial attention mechanism only at the lowest scale $S$ for efficiency, while using skip connections to preserve high-frequency information. The resulting feature maps are then decoded to predict per-pixel depth $\hat{\mathbf{D}}_{t} \in \mathbb{R}^{N \times H \times W}$ for each of the $N$ images.

To train our model (see Sec.~\ref{sec:train}), the depth network takes the target frame $\mathbf{I}{_{t}}$ and predicts a depth map for each of the $N$ images. The network is supervised based on the spatial photometric consistency between the target images in $\mathbf{I}{_{t}}$. However, since the spatial overlap between images in such a setup is typically minimal, we additionally supervise our model temporally. For that, a pose network takes the front view images from the target frame $\mathbf{I}{_{t}}$ and from a source frame $\mathbf{I}{_{t'}}$, where $t'$ is either a past frame $t-1$ or a future frame $t+1$, and predicts the transformation of the camera poses between $t$ and $t'$. This transformation is used to re-render the target frame from the source frame to enforce temporal photometric consistency.

\subsection{Multi-View Consistency}
In a multi-view setup, processing each image in isolation can yield inconsistent depth predictions across the images, i.e., the same point in 3D object space observed in multiple images may be predicted to be at different 3D locations for each image, since the individual feature representations are unshared and image‑specific. To address this issue, we propose an explicit geometry-guided enforcement of multi-view consistency. We first project the pixel coordinates of all individual feature maps onto a shared unit cylinder. This results in a cylindrical position map $ \mathbf{O}{_{S, \mathbf{I}_{t,i}}} \in\mathbb{R}^{H_S\times W_S\times 2}$ for each image $\mathbf{I}_{t, i}$, where $i \in N$. Attention between pixels is then applied with respect to their cylindrical distance.
\paragraph{Cylindrical Projection}
\label{cylproj}
We project the pixel positions of the image features $\mathbf{F}{_{S, \mathbf{I}_t}}$ of the lowest spatial resolution $S$, extracted by the encoder and originally given in the respective image coordinate system, onto a common unit cylinder. This cylindrical projection produces a unified representation, i.e., the information from all images is transformed into a common coordinate system. A cylindrical representation is well suited to surround camera setups, yielding a circular topology in which views wrap around, and every image connects to its neighbors, while avoiding the pole-related distortions of spherical representations. However, the conventional approach to cylindrical image stitching assumes that each pair of overlapping images can be related by a single homography. In surround camera setups, this assumption is often violated due to the non-negligible baselines between the cameras.  Applying such methods to images with a significant baseline induces parallax, whereby the same scene elements project to different locations on the cylinder, leading to misalignment (e.g., ghosting effect).

\begin{figure}[t] 
    \centering
    \includegraphics[width=0.34\textwidth]{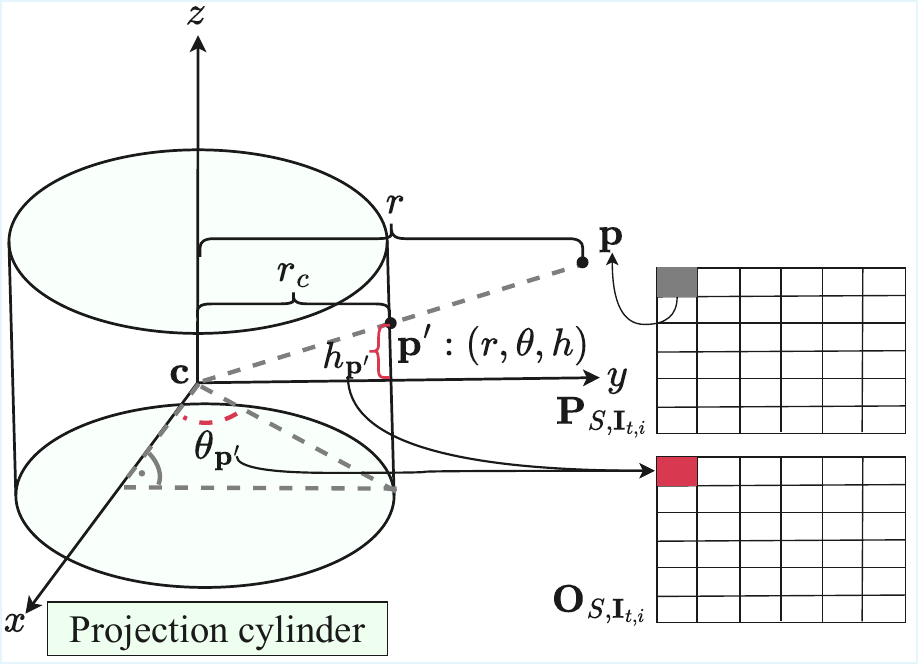}
    \caption{\normalsize Visualization of the cylindrical projection of a pixel $p$ from the 3D position map $\mathbf{P}{_{S, \mathbf{I}_{t, i}}}$ resulting in cylindrical position map $\mathbf{O}{_{S, \mathbf{I}_{t, i}}}$ for all pixels in $\mathbf{P}{_{S, \mathbf{I}_{t, i}}}$.}
    \label{fig:cylinderproj}
\end{figure}

Thus, we first reconstruct the scene in 3D space, using the  preliminary depth map predicted for each image separately by our depth network. The resulting 3D points are then projected onto a unit-radius cylinder. For a feature map $\mathbf{F}{_{S, \mathbf{I}_{t, i}}} \in\mathbb{R}^{H_S\times W_S\times F_S}$ of image $\mathbf{I}_{t, i}$ given its intrinsics $\mathbf{K}{_{\mathbf{I}_{t, i}}} \in\mathbb{R}^{3\times3}$, its pose relative to a common reference coordinate system on the rig ${}^{\text{ref}}\!\mathbf{T}_{\mathbf{I}_{t, i}}\in\mathbb{R}^{4\times4}$, and a preliminarily estimated depth $\mathbf{\hat{D}}{_{{\mathbf{I}_{t, i}}}}$, we back‑project the pixels to 3D to obtain a 3D position map $ \mathbf{P}{_{S, \mathbf{I}_{t, i}}} \in\mathbb{R}^{H_S\times W_S\times 3}$:
\begin{align}
   \mathbf{P}{_{S, \mathbf{I}_{t, i}}} = \Pi(\mathbf{F}{_{S, \mathbf{I}_{t, i}}}, \mathbf{K}{_{\mathbf{I}_{t, i}}}, {}^{\text{ref}}\!\mathbf{T}_{\mathbf{I}_{t, i}}, \mathbf{\hat{D}}{_{{\mathbf{I}_{t, i}}}})\,,
\end{align}
where $\Pi$ is the mapping from 2D to 3D. Let $\mathbf{p} \in \mathbb{R}^3$ be a single 3D point obtained from $\mathbf{P}{_{S, \mathbf{I}_{t, i}}}$. We fix a unit cylinder with radius $r_c=1$ and center $\mathbf{c} = (x_c, y_c, z_c)$, with its central axis being parallel to the $z$-axis. The distance in the xy-plane between $\mathbf{p_o}\;=\; \mathbf{p}-\mathbf{c} \;=\; (x_o, y_o, z_o)$ and the cylinder's vertical axis through $\mathbf{c}$ is defined as $ r= \sqrt{{x_o}^2+{y_o}^2}$. We project $\mathbf{p_o}$ onto the lateral surface of the cylinder via a central projection with the projection center located in $\mathbf{c}$ (see Fig.~\ref{fig:cylinderproj}). Consider the ray $\ell(b)=\mathbf{c}+b\,\mathbf{p_o}$ for $b\in\mathbb{R}$. 
The intersection with the cylinder's surface $\mathcal{C}=\{\,\mathbf{q}\in\mathbb{R}^3 | \| (\mathbf{q}-\mathbf{c})_{xy}\| = r_c \,\}$, with $(\mathbf{q}-\mathbf{c})_{xy}$ denoting the projection onto the $xy$-plane,
is given by:
\begin{align}
\| ( \ell(b)-\mathbf{c} )_{xy} \| \;=\; \| (b\,\mathbf{p_o})_{xy} \| \;=\; |b|\,r_c \;=\; r,
\label{eq:3}
\end{align}
Based on Eq.~\ref{eq:3}, for $\mathbf{p_o}$, the projected point $\mathbf{p'}=(x',y',z')$ on the cylinder is given as:
\begin{align}
\mathbf{p'} \;=\; \mathbf{c} + b\,\mathbf{p_o} \;=\; \mathbf{c} - \frac{r}{r_c}\,\mathbf{p_o} .
\end{align}
We then parameterize $\mathbf{p'}$ in cylindrical coordinates by its azimuth $\theta_{\mathbf{p'}}$ and height $h_{\mathbf{p'}}$:
\begin{align}
\theta_{\mathbf{p'}} &= \operatorname{atan2}(y'-y_c,\;x'-x_c) \in (-\pi,\pi], \\
h_{\mathbf{p'}} &= z'-z_c .
\end{align}
For each feature map $\mathbf{F}{_{S, \mathbf{I}_{t, i}}}$, we obtain an associated position map $ \mathbf{O}{_{S, \mathbf{I}_{t, i}}} \in\mathbb{R}^{H_S\times W_S\times 2}$ that encodes the pixel positions on the unit cylinder by the azimuth angle and height.

\paragraph{Spatial Attention}
We adopt this cylindrical representation as it maps corresponding pixels from different images into nearby locations on the cylinder, even when the initial depth predictions are inaccurate.
In contrast, operating directly in 3D would cause corresponding pixels or nearby pixels to be mapped far apart if the initial depth predictions are inaccurate. Based on the spatial proximity, we enable the exchange of features between pixels and across images using a novel non-learned attention weighting. We define the attention weights based on the geodesic distance between the pixels on the cylinder. 
This approach allows us to incorporate the geometric relation between the images into our attention mechanism, particularly at inference time. Thus, it enables pixels to exchange contextual features in a way that respects the geometric relation between their corresponding 3D object points, thereby promoting depth predictions that are consistent across images.
In contrast, purely learned attention does not inherently exploit the known geometric relationships between images.

We model the spatial attention weights using a truncated 2D Gaussian kernel centered at a query pixel on the cylinder. We assume that spatially close pixels in 3D lie within a local neighborhood on the cylinder; the Gaussian provides a soft weighting to account for minor errors in the projection as well. The truncation of the Gaussian is important to avoid the consideration of feature information of distant and thus irrelevant pixels. The spatial attention weight $a^{sp}_{uv}$ for a pixel pair $u,v$ from $\mathbf{F}{_{S, \mathbf{I}_t}}$, and their positions  on the cylinder $\mathbf{o}{_u}$ and $\mathbf{o}{_v}$ from $\mathbf{O}{_{S, \mathbf{I}_t}}$, is given as:
\begin{align}
d_{ij}^2 &= (d_{geo}( \mathbf{o}_{i}, \mathbf{o}_j))^{\top}\mathbf{\Sigma}^{-1}\,d_{geo}(\mathbf{o}_{i}, \mathbf{o}_j), \label{eq:dij}\\
a^{sp}_{ij} &=
\begin{cases}
\exp\!\left(-\tfrac{1}{2}\,d_{ij}^2\right), & d_{ij}^2 \le \tau^2,\\
0, & \text{otherwise},
\end{cases} \label{eq:asij}
\end{align}
where $\mathbf{\Sigma}$ is a pre-defined non-learned covariance matrix defining the shape and size of the 2D Gaussian kernel, $\tau$ is the truncation threshold and $d_{geo}$ is the geodesic distance. The feature vector for a pixel $u$ modulated by the attention weights, for all possible pixels of $v$, is given as:

\begin{equation}
\begin{aligned}
\mathbf{f}_{u}' &= \sum_{v} a^{sp}_{uv} \mathbf{W}_V \mathbf{f}_v,
\end{aligned}
\label{eq:attention}
\end{equation}where $\mathbf{W}_V$ is a learned projection matrix. For all pixels in  $\mathbf{F}{_{S, \mathbf{I}_t}}$, the resulting feature maps modulated by the attention are given as $\mathbf{F'}{_{S, \mathbf{I}_t}} \in\mathbb{R}^{N\times H_S\times W_S\times F_S}$. 
The final depth $\hat{\mathbf{D}}_{t}$ is produced by feeding $\mathbf{F'}_{S,\mathbf{I}_t}$ and the feature maps $\mathbf{F}_{s, \mathbf{I}_t}$ from all scales except $S$ into the decoder.

\subsection{Self-Supervision}
\label{sec:train}
Our method is trained in a self-supervised manner, enforcing photometric consistency between images. The photometric loss~\cite{godard2017unsupervised} compares a target image $\mathbf{{I}}{{_{t, i}}}\in\mathbb{R}^{ H\times W\times 3}$ with a re-rendered target image $\mathbf{\hat{I}}_{t, i}$ from the source images $\mathbf{I}_{\{t, t'\}}$ and is defined as:
\begin{align}
\mathcal{L}_{photo} &= \frac{1}{M} \sum_{M} 
\alpha \frac{1 - \text{SSIM}(\mathbf{\hat{I}}_{t,i}, \mathbf{I}_{t, i})}{2} \notag \\
&\quad + (1 - \alpha) \left\| \mathbf{\hat{I}}_{t,i} - \mathbf{I}_{t,i} \right\|. 
\label{Eq:photo}
\end{align}
where $\alpha = 0.85$, SSIM~\cite{wang2004image} is the structural similarity, and $M=H\cdot W$ is the number of pixels in the image.
The rendering can either be done temporally, between images from two consecutive frames, spatially, between different cameras on the rig, or spatio-temporally as a combination of both. These three configurations result in three variants of the photometric loss, described in more detail in the following.
Our overall loss is defined as the weighted sum of these photometric loss terms and a set of auxiliary losses:
\begin{align}
\mathcal{L} &= \mathcal{L}_{\text{photo,temp}}
+ \lambda_{sp}\mathcal{L}_{\text{photo,sp}}
+ \lambda_{spt}\mathcal{L}_{\text{photo,spt}} \nonumber \\
&\quad
+ \lambda_{sm}\mathcal{L}_{sm}
+ \lambda_{DCCL}\mathcal{L}_{DCCL}
+ \lambda_{MVRCL}\mathcal{L}_{MVRCL},
\label{Eq:loss}
\end{align}
where $\mathcal{L}_{sm}$ is an edge-aware smoothing loss of the depth~\cite{godard2017unsupervised}, $\mathcal{L}_{DCCL}$~\cite{ding2024towards} is a dense depth consistency loss that enforces consistency of the depth predictions between spatially adjacent images, and $\mathcal{L}_{MVRCL}$~\cite{ding2024towards} enforces photometric consistency of the spatial and spatio-temporal reconstructions. $\lambda$ are weighting factors.

\paragraph{Spatial Loss}

Given the metric relative poses, we make use of the spatial overlap between images from the same frame to obtain a supervision signal based on stereo matching. This enables the network to predict depth that is consistent in scale and given in metric units in the overlapping regions and, due to the propagation of information, also beyond. In our work, to better address holes, we employ inverse warping~\cite{godard2019digging}: each pixel $\mathbf{p}_{\mathbf{I}_{t,i}}$ in a target image $\mathbf{I}_{t,i}$ is projected into the coordinate system of a spatially adjacent source image $\mathbf{I}_{t,j}$ using the predicted depth $\mathbf{\hat{D}}{_{\mathbf{I}_{t,i}}}$ and the metric relative pose ${}^{\mathbf{I}_{t,j}}\!\mathbf{T}_{\mathbf{I}_{t,i}} ^ {}$ between these images:

\begin{align}
    \mathbf{\hat{p}}_{\mathbf{I}_{t,j}}=\mathbf{K}{_{\mathbf{I}_{t, j}} {}^{\mathbf{I}_{t,j}}\!\mathbf{T}_{\mathbf{I}_{t,i}} ^ {} \mathbf{\hat{D}}{_{\mathbf{I}_{t,i}}} \textbf{K}_{\mathbf{I}_{t, i}}^{-1}} \mathbf{p}_{\mathbf{I}_{t,i}}.
    \label{eq:11}
\end{align}
A new target image is rendered by sampling from the source image according to Eq.~\ref{eq:11}. The spatial loss $\mathcal{L}_{photo, sp}$ is then defined as the photometric loss (Eq.~\ref{Eq:photo}) between the target image and the re-rendered target image.

\begin{figure*}
    \centering
    \includegraphics[width=0.81\textwidth]{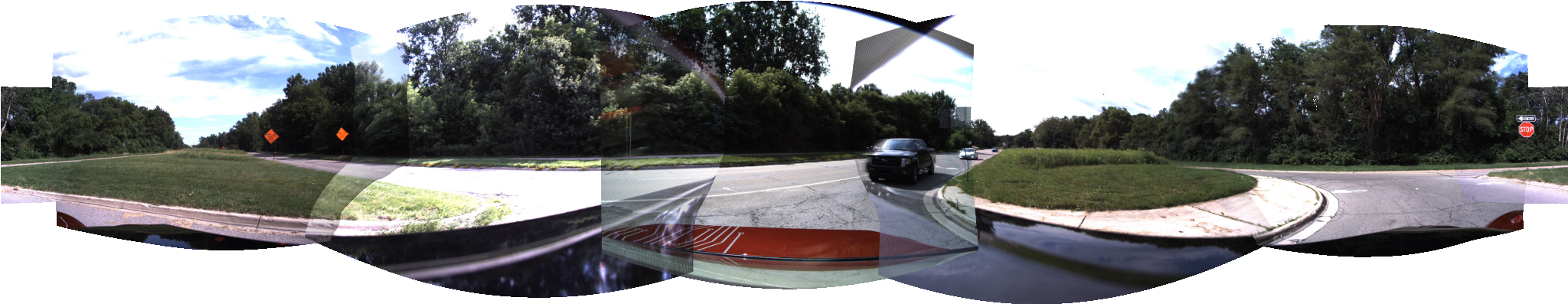}
    \caption{\normalsize Panoramic visualization of the cylindrical projection of RGB inputs. Note that in our method, only pixel positions are projected, not RGB values. This figure is provided solely for illustration, to show how objects captured from different views are mapped to nearby locations in cylindrical coordinates.}
    \label{fig:panorama}
\end{figure*}

\begin{figure}[t]
    \centering
    \begin{subfigure}[t]{0.495\linewidth}
        \centering
        \includegraphics[width=\linewidth]{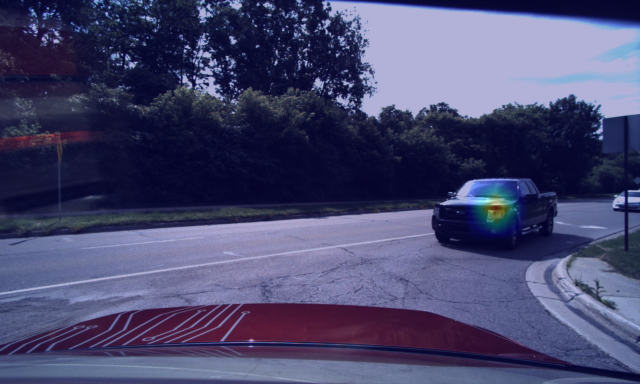}
        \subcaption{Back image}
        \label{fig:two-right}
    \end{subfigure}\hfill
    \begin{subfigure}[t]{0.495\linewidth}
        \centering
        \includegraphics[width=\linewidth]{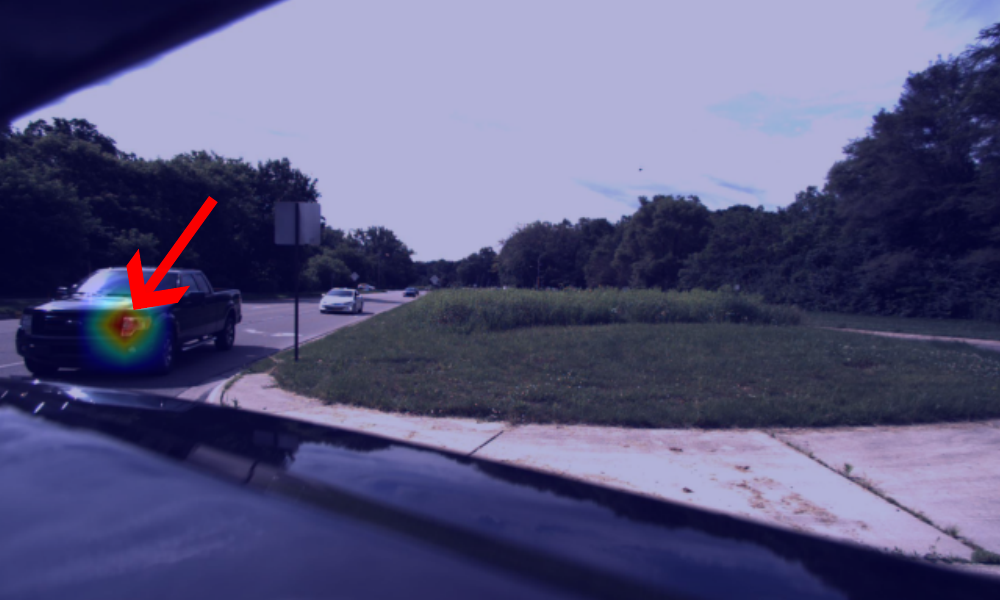}
        \subcaption{Back-left image}
        \label{fig:two-left}
    \end{subfigure}\hfill
    \caption{\normalsize Attention maps for a query token (indicated by the arrow in the back-left image), as overlays on the respective RGB images, showing that this token attends to itself, nearby regions, and to the corresponding region in the spatially adjacent image. High attention is shown in red, low attention in yellow to blue.}
    \label{fig:attn}
\end{figure}

\paragraph{Temporal Loss}

Due to the limited spatial overlap between images from the same frame, spatial supervision alone is insufficient for learning accurate depth estimation. To address this limitation, we use temporal context by enforcing photometric consistency between $\mathbf{I}_{t,i}$ and its temporally adjacent source image $\mathbf{I}_{t',i}$, based on a predicted pose between two frames ${}^{\mathbf{I}_{t',i}}\!\mathbf{\hat{T}}_{\mathbf{I}_{t,i}} ^ {}$.
The temporal loss $\mathcal{L}_{photo, temp}$ is given as the photometric loss (Eq.~\ref{Eq:photo}) between a target image and a re-rendered target image from a temporal source image. To estimate the pose across time, we assume that all cameras share the same motion, i.e., that they are mounted rigidly to each other. Following~\cite{ding2024towards}, we use only the front image to predict the front-camera temporal pose ${}^{\mathbf{I}_{t',1}}\!\hat{\mathbf{T}}_{\mathbf{I}_{t,1}}$ using the pose network, ensuring lightweight computations. The pose ${}^{\mathbf{I}_{t',i}}\!\mathbf{\hat{T}}_{\mathbf{I}_{t,i}} ^ {} =  {{}^{\mathbf{I}_{t,1}}\!\mathbf{T}^{-1}_{\mathbf{I}_{t,i}}}{}^{\mathbf{I}_{t',1}}\!\hat{\mathbf{T}}_{\mathbf{I}_{t,1}}{}^{\mathbf{I}_{t,1}}\!\mathbf{T}_{\mathbf{I}_{t,i}}$ is derived based on the given camera pose w.r.t the front camera ${}^{\mathbf{I}_{t,1}}\!\mathbf{T}_{\mathbf{I}_{t,i}}$. 

\paragraph{Spatio-Temporal Loss}
Following~\cite{guizilini2022full}, we employ a spatio-temporal loss, enforcing photometric consistency between images taken by different cameras and at different points in time. This allows us to further increase the number of object points that are seen in more than one image and, thus, to better learn metric scale. The warping follows the same principle as in the previous losses, where a new target image $\mathbf{I}_{t,i}$ is rendered from a source image $\mathbf{I}_{t',j}$ based on the spatio-temporal pose ${}^{\mathbf{I}_{t',j}}\!\mathbf{\hat{T}}_{\mathbf{I}_{t,i}} ^ {} = {}^{\mathbf{I}_{t',j}}\!\mathbf{\hat{T}}_{\mathbf{I}_{t,j}} ^ {} {}^{\mathbf{I}_{t,j}}\!\mathbf{T}_{\mathbf{I}_{t,i}} ^ {}$. The spatio-temporal loss $\mathcal{L}_{\text{photo},\,\text{spt}}$ is defined as the photometric loss (Eq.~\ref{Eq:photo}) between the target image and a re-rendered target image from a spatio-temporal source image.

%% file: sec/4_experiments.tex
\section{Experiments}

\subsection{Experimental Setup}
\label{sec:expsetup}

\begin{table}[!tbp]
\centering

\label{tab:ddad_nuscenes}
\setlength{\tabcolsep}{3pt}
\renewcommand{\arraystretch}{1.05}
\scriptsize
\resizebox{\columnwidth}{!}{%
\begin{tabular}{@{}llcccc@{}}
\toprule
Dataset & Method &
Abs Rel & Sq Rel [m] & RMSE [m] & $\delta < 1.25$ \\
\midrule
\multirow{6}{*}{\textbf{DDAD}} 
& FSM                & \textbf{0.201} &  - & - & - \\
& FSM*                & 0.228 &  4.409 & 13.43 & 68.7 \\
& VFDepth            & 0.218 &  3.660 & 13.32 & 67.4 \\
& SurroundDepth      & 0.208 &  \textbf{3.371} & 12.97 &  69.3 \\
& CVCDepth         & 0.210 &  3.458 & 12.87 &  70.4 \\
& \textbf{CylinderDepth (ours)} & 0.207 & 3.503 & \textbf{12.76} & \textbf{70.9} \\
\midrule
\multirow{6}{*}{\textbf{nuScenes}}
& FSM                & 0.297 &  - &  - &  - \\
& FSM*                & 0.319 &  7.534 & 7.86 & 71.6 \\
& VFDepth            & 0.289 &   5.718 & 7.55 &  70.9 \\
& SurroundDepth      & 0.280 &  \textbf{4.401} & 7.46 & 66.1 \\
& CVCDepth         &  0.264 &  5.525 &   7.17 &  76.3 \\
& \textbf{CylinderDepth (ours)} & \textbf{0.244} & 6.025 &  \textbf{6.82} & \textbf{80.5} \\
\bottomrule
\end{tabular}%
}
\caption{Comparison of our method with state-of-the-art methods. FSM* denotes results reproduced with the implementation of \cite{kim2022self}. $\delta$ is given in [\%]. Abs Rel is unit-free.}
\label{tab:method_compar}
\end{table}

\begin{table}[!tbp]
\centering
\setlength{\tabcolsep}{5pt} 
\renewcommand{\arraystretch}{1.03}
\scriptsize

\resizebox{0.95\columnwidth}{!}{%
\begin{tabular}{@{}llcc@{}}
\toprule
Dataset & Method & Abs Rel & Depth Cons [m] \\
\midrule
\multirow{4}{*}{\textbf{DDAD}} 
& VFDepth (3D) & 0.222 & \textbf{4.82} \\
\cmidrule(lr){2-4}
& SurroundDepth (2D) & 0.217 & 7.86 \\
& CVCDepth (2D) & 0.212 & 6.35 \\
& \textbf{CylinderDepth (ours) (2D)} & \textbf{0.207} & \textbf{5.68} \\
\midrule
\multirow{4}{*}{\textbf{nuScenes}}
& VFDepth (3D) & \textbf{0.277} & \textbf{3.57} \\
\cmidrule(lr){2-4}
& SurroundDepth (2D) & 0.295 & 6.33 \\
& CVCDepth (2D) & 0.388 & 3.02 \\
& \textbf{CylinderDepth (ours) (2D)} & \textbf{0.218} & \textbf{2.69} \\
\bottomrule
\end{tabular}
}
\caption{Comparison of our method with state-of-the-art 2D and 3D methods in overlapping regions. The best results per category are shown in bold. Abs Rel is unit-free.}
\label{tab:overlap}
\end{table}

\paragraph{Dataset}
We train and evaluate our method on DDAD ~\cite{guizilini20203d} and nuScenes~\cite{caesar2020nuscenes}. Both datasets provide images from a six-camera surround rig mounted on a vehicle, capturing 360$^\circ$ of the vehicle's surrounding, along with LiDAR-derived reference depth. We resize the images to 384$\times$640 pixels for DDAD and 352$\times$640 pixels for nuScenes before providing them as input to our model. Depth is evaluated up to 200\,m for DDAD and 80\,m for nuScenes, corresponding to the range of the ground-truth depth labels. Following \cite{wei2023surrounddepth, kim2022self}, we apply self-occlusion masks for DDAD to remove the ego-vehicle from the images during training.

\paragraph{Implementation Details}
We use a ResNet-18~\cite{he2016deep} encoder pre-trained on ImageNet~\cite{deng2009imagenet} for the depth and pose networks. The decoder in both networks is adopted from~\cite{godard2019digging} and is randomly initialized. Training is conducted on 8 NVIDIA RTX 3060 GPUs with a batch size of 1 (consisting of six surround images) per GPU. We optimize the network using Adam~\cite{kingma2014adam} with
$\beta_{1}=0.9$ and $\beta_{2}=0.999$. The initial learning rate is $10^{-4}$ with a StepLR scheduler decreasing the learning rate by a factor of $0.1$ after completing $\tfrac{3}{4}$ of the total 20 training epochs. For the Gaussian distribution in Eq.~\ref{eq:dij}, we use a covariance matrix $\mathbf{\Sigma} = \mathrm{diag}(0.02,\,0.02)$, and $\tau = 1.2$. These values are selected based on the feature-map resolution. For the hyperparameter in Eq.~\ref{Eq:loss}, we choose $\lambda_{sp}=0.03 $, $\lambda_{spt}=0.1$, $\lambda_{sm}=0.1$, $\lambda_{DCCL}=1\times 10^{-3}$ and  $\lambda_{MVRCL}=0.2$ based on preliminary experiments.

\begin{figure*}[t]
\centering
\setlength{\tabcolsep}{1pt}
\renewcommand{\arraystretch}{0.0}
\begin{tabular}{*{5}{c}}
\scriptsize Image & \scriptsize VFDepth & \scriptsize SurroundDepth & \scriptsize  CVCDepth & \scriptsize  Ours \\

\includegraphics[width=0.19\textwidth]{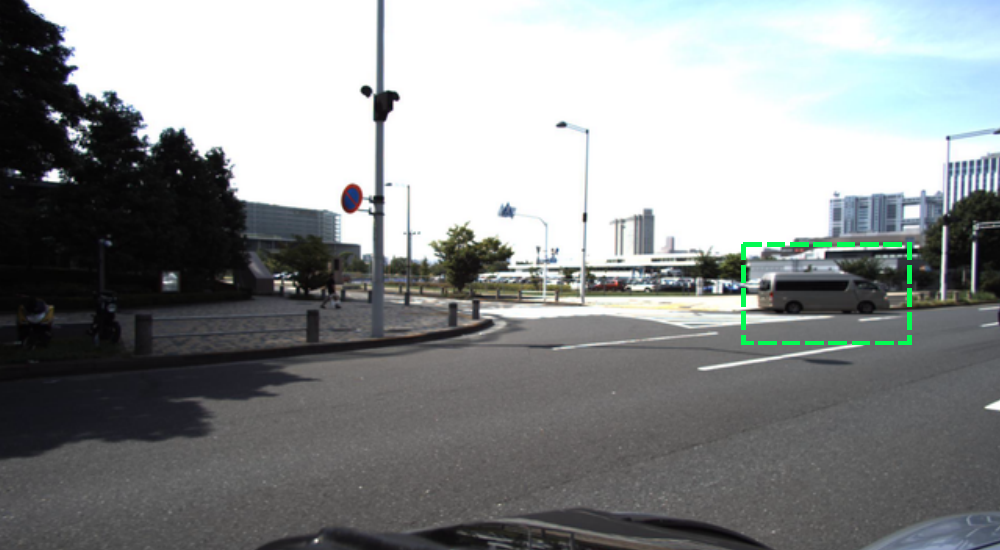} &
\includegraphics[width=0.19\textwidth]{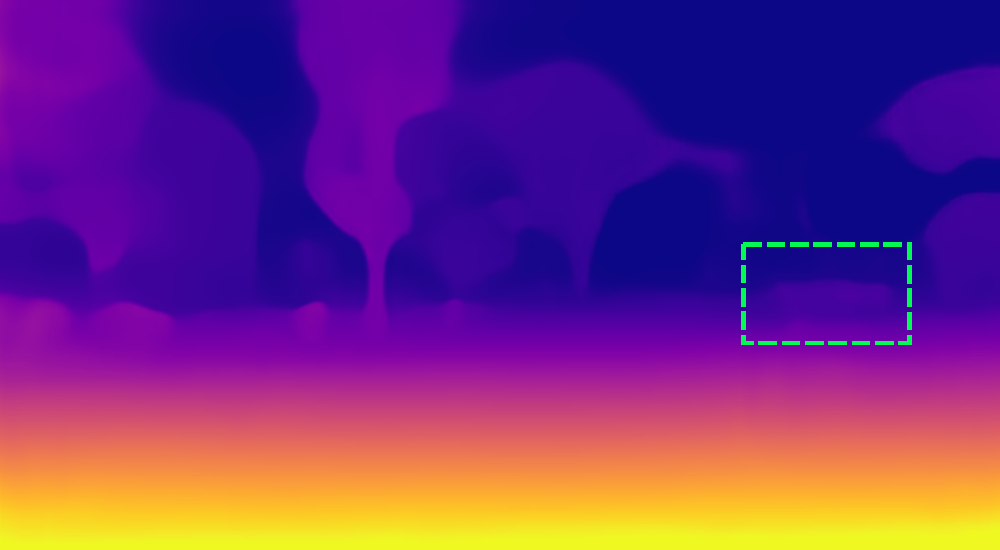} &
\includegraphics[width=0.19\textwidth]{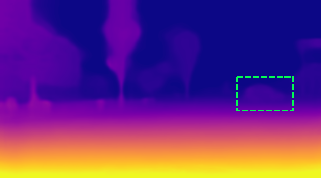} &
\includegraphics[width=0.19\textwidth]{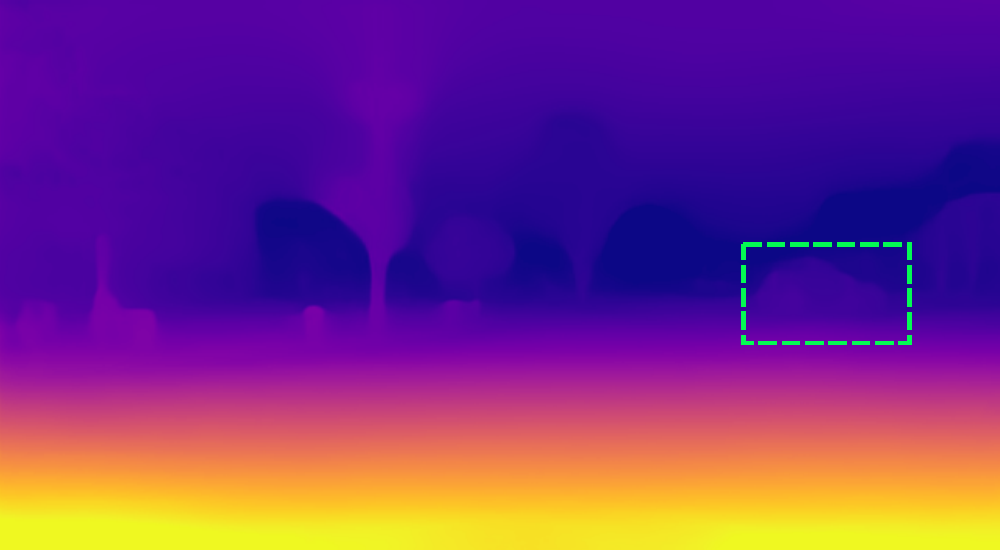} &
\includegraphics[width=0.19\textwidth]{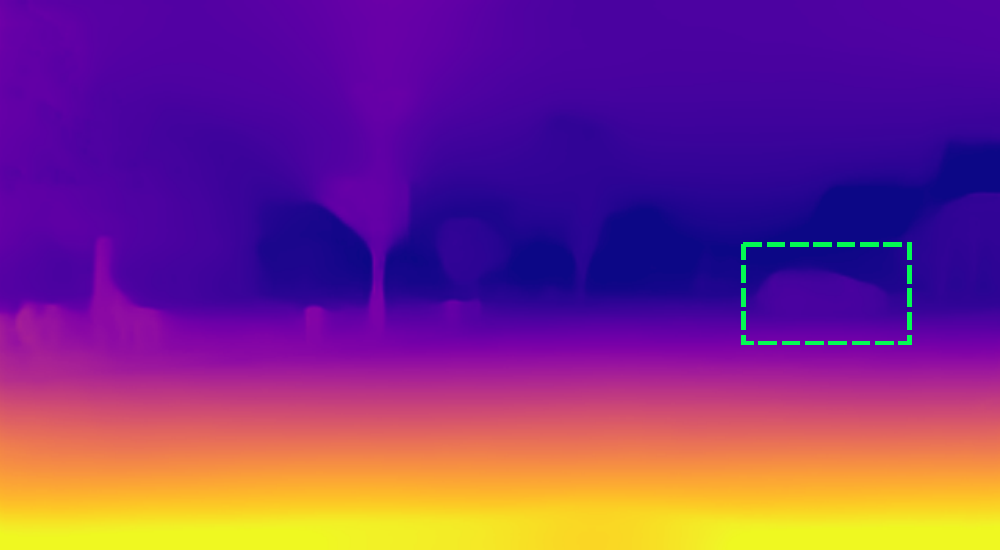} \\[2pt]

\includegraphics[width=0.19\textwidth]{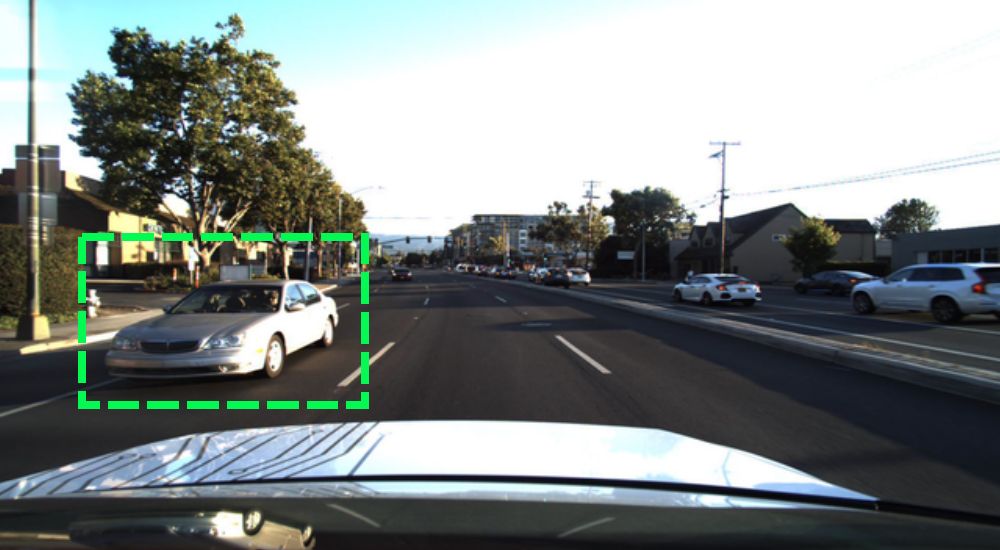} &
\includegraphics[width=0.19\textwidth]{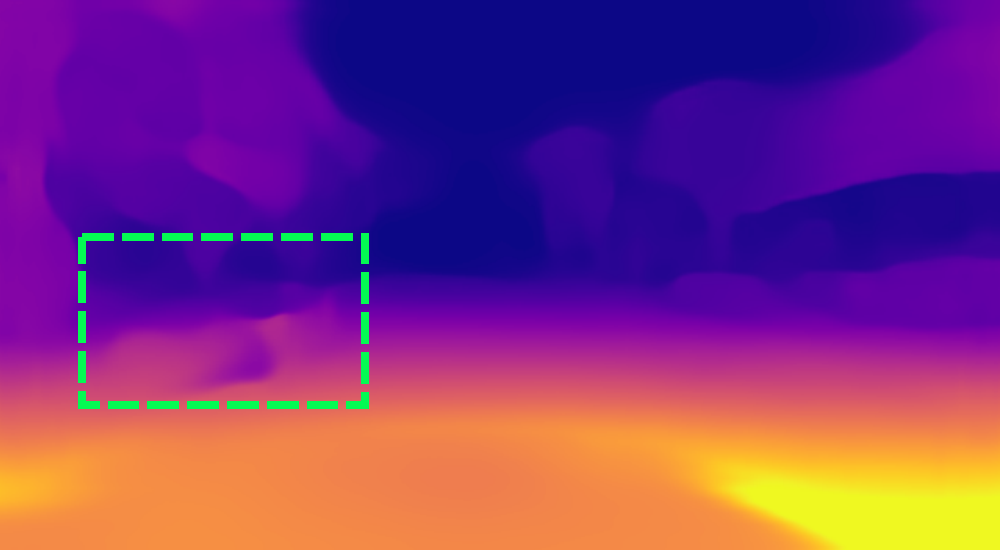} &
\includegraphics[width=0.19\textwidth]{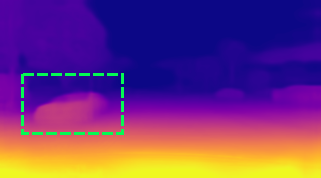} &
\includegraphics[width=0.19\textwidth]{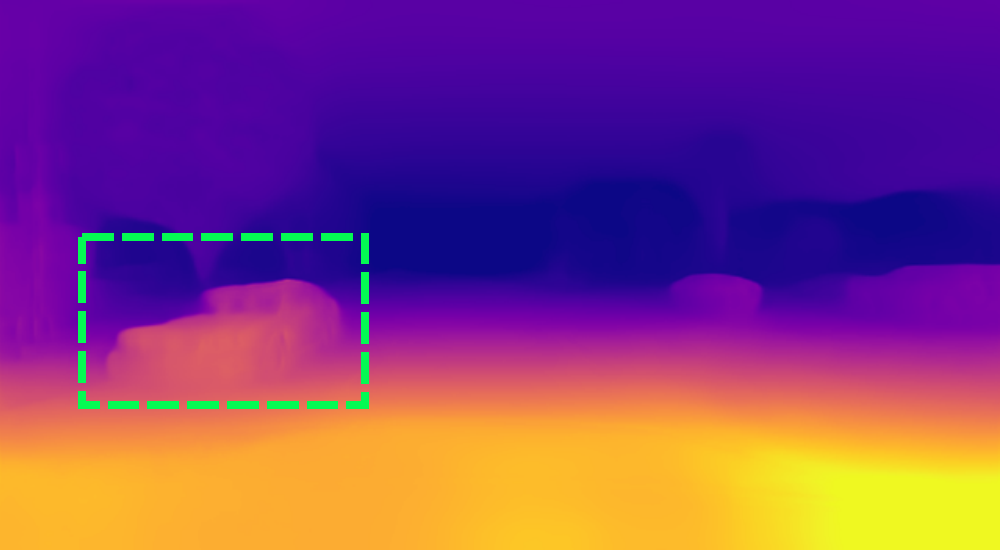} &
\includegraphics[width=0.19\textwidth]{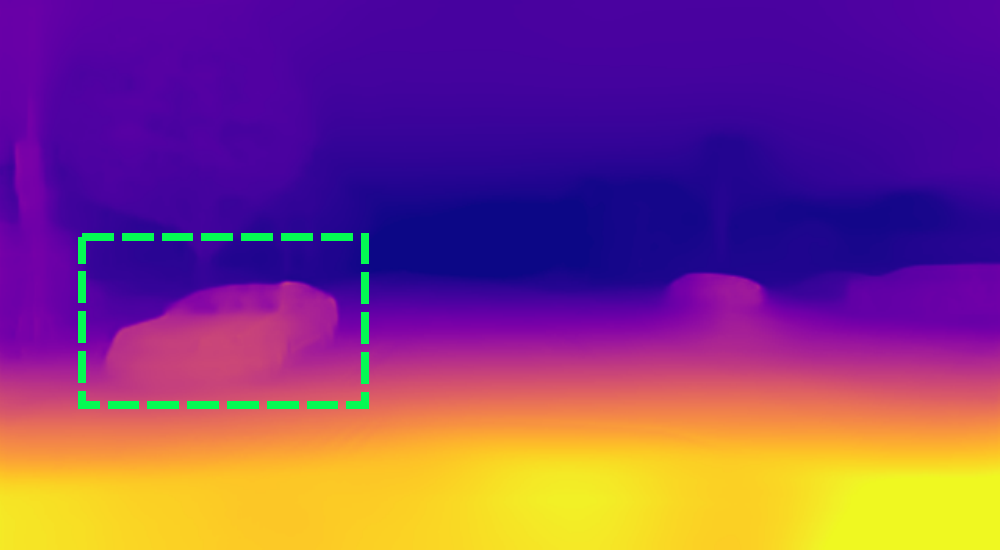} \\
\end{tabular}

\caption{Comparison of depth maps predicted by our method and by state-of-the-art methods on DDAD. Our results show better preserved details and well-defined object boundaries (green bounding boxes). Depth is shown from close in yellow to distant in blue.}
\label{fig:method_comparison}
\end{figure*}
\paragraph{Evaluation Metrics}
We adopt standard depth evaluation metrics~\cite{eigen2014depth}: Absolute relative difference (Abs Rel), Squared Relative difference (Sq Rel), RMSE, and the percentage of pixels with an error below a threshold $\delta$. In addition, we propose a novel quality metric to assess the multi-view depth consistency (Depth Cons): For each pair of corresponding pixels in the overlapping regions, the depth value of each pixel is converted into a Euclidean distance from a common reference coordinate system. The RMSE is then computed between the Euclidean distances of the pixel and its correspondence (see supp. material).

\subsection{Experimental Results}

\begin{figure*}[t]
\centering
\begin{subfigure}{0.312\linewidth}
    \centering
    \includegraphics[width=\linewidth]{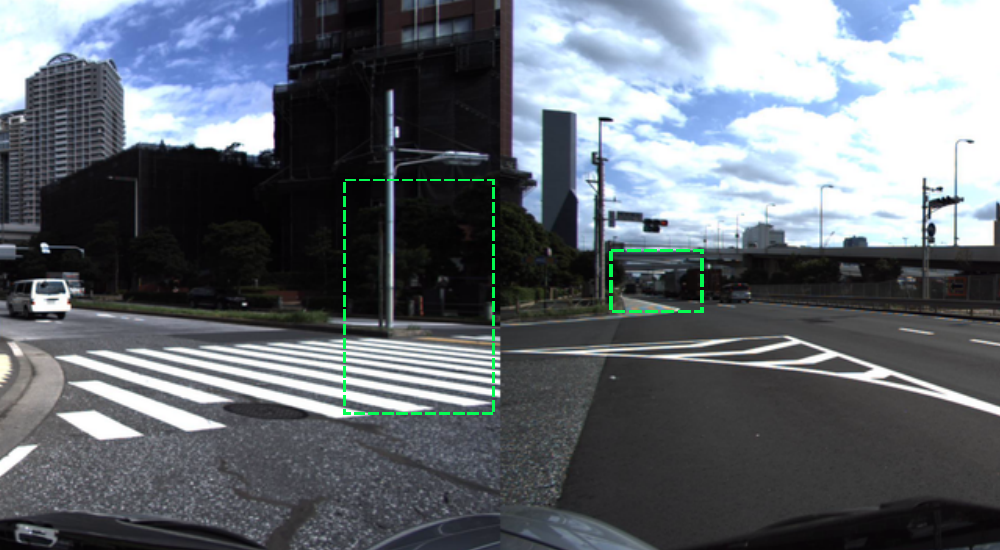}
    \caption{combined front-right and front-left image}
\end{subfigure}
\begin{subfigure}{0.312\linewidth}
    \centering
    \includegraphics[width=\linewidth]{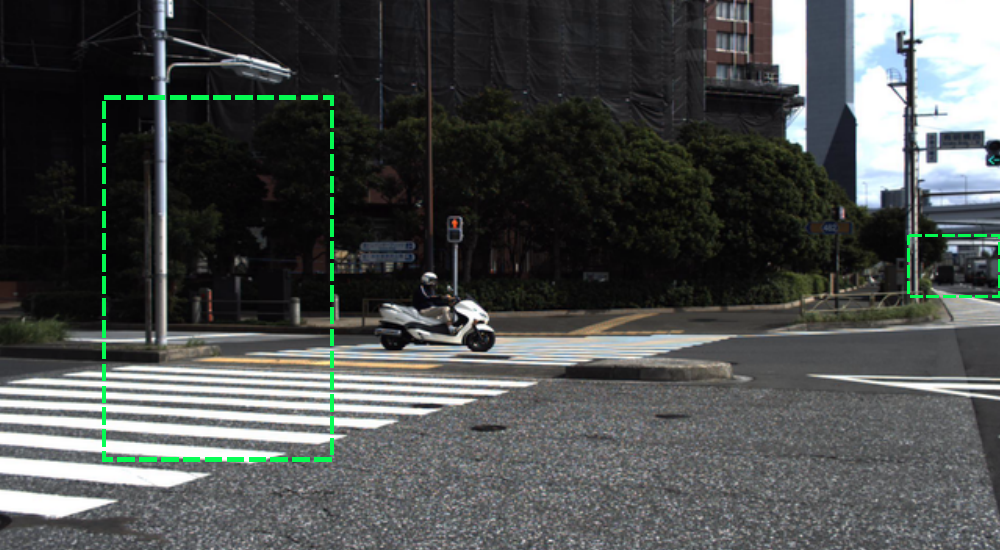}
    \caption{Front image}
\end{subfigure}
\begin{subfigure}{0.312\linewidth}
    \centering
    \includegraphics[width=\linewidth]{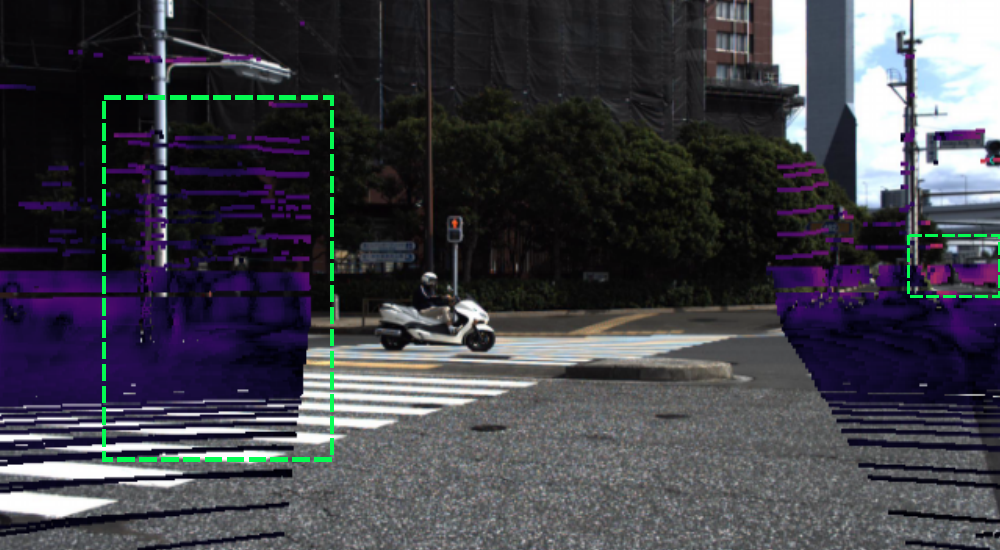}
    \caption{CylinderDepth (ours)}
\end{subfigure}

\begin{subfigure}{0.312\linewidth}
    \centering
    \includegraphics[width=\linewidth]{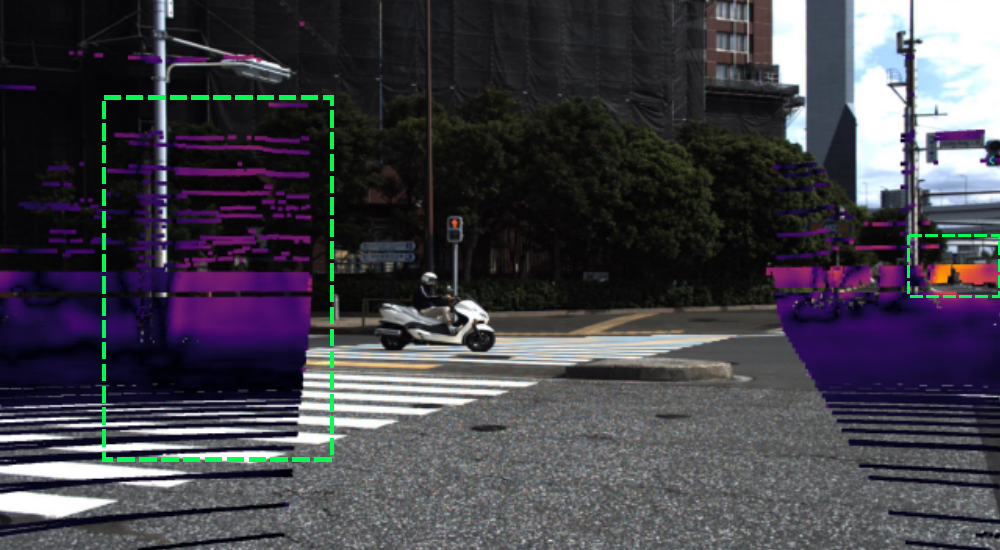}
    \caption{SurroundDepth}
\end{subfigure}
\begin{subfigure}{0.312\linewidth}
    \centering
    \includegraphics[width=\linewidth]{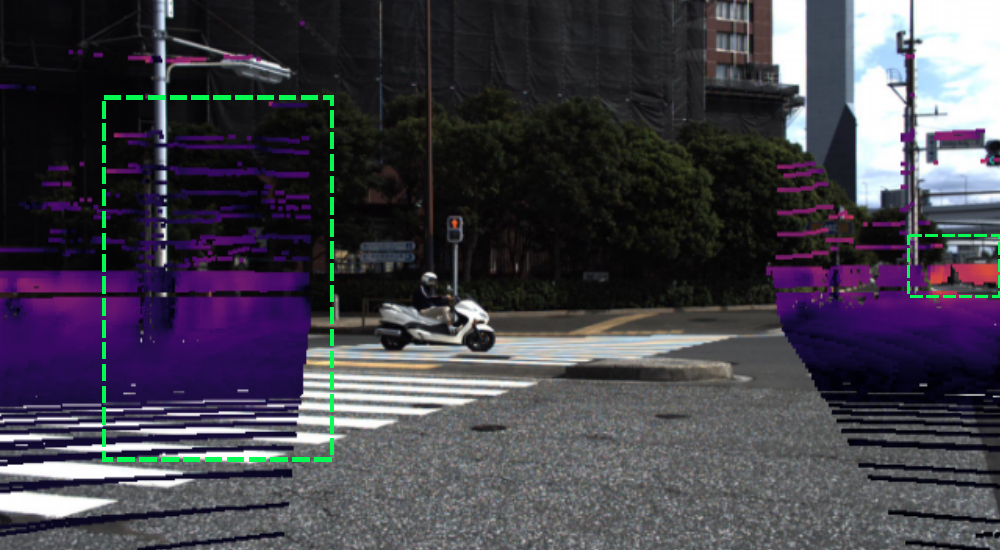}
    \caption{CVCDepth}
\end{subfigure}
\begin{subfigure}{0.312\linewidth}
    \centering
    \includegraphics[width=\linewidth]{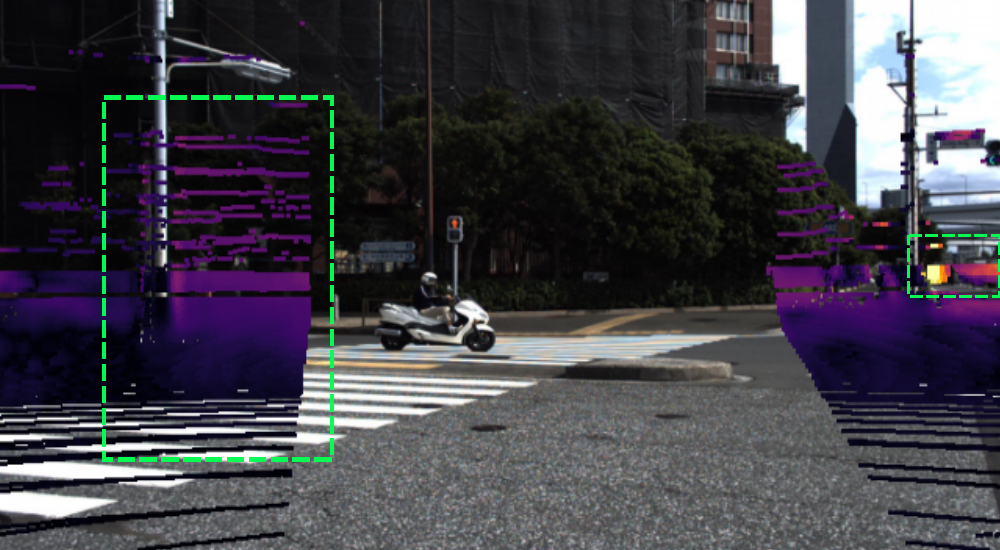}	
    \caption{VFDepth}
\end{subfigure}

\caption{Exemplary depth consistency error maps, computed using the metric described in Sec.~\ref{sec:expsetup} overlayed on the front-image, comparing our approach with state-of-the-art methods on DDAD. Our method maps overlapping regions from the two images to nearby 3D coordinates. In contrast, the other methods exhibit a higher inconsistency in these regions (green bounding boxes). (a) shows the combined relevant regions from the front-right and front-left images that overlap with the front image. Errors are visualized using the inferno colormap, ranging from black (low error) to yellow (high error).}
\label{fig:3d}
\end{figure*}

We compare our method against four state-of-the-art methods: FSM~\cite{guizilini2022full}, SurroundDepth~\cite{wei2023surrounddepth}, VFDepth~\cite{kim2022self}, and CVCDepth~\cite{ding2024towards}. Since the code of FSM is not publicly available, we report the related results from the original paper and as reproduced in \cite{kim2022self}. For CVCDepth, we compare against their ResNet18 version. As shown in Fig.~\ref{fig:method_comparison} and~\ref{fig:3d} and Tab.~\ref{tab:method_compar} and~\ref{tab:overlap}, our approach achieves substantial improvements in multi-view depth consistency over other 2D-based and 3D-based depth estimation methods~\cite{wei2023surrounddepth, ding2024towards}.

Our method also achieves slightly higher depth accuracy in both, overlapping regions and full-image evaluations on both datasets. However, it is to be noted that in nuScenes, cameras are synchronized with the LiDAR sweep, leading to clear time differences between images captured by different cameras (up to 40\,ms). In dynamic scenes and under rig motion, larger deviations from the time-synchronization assumption degrade result quality. This issue affects all methods modeling shared camera motion and relying on spatial supervision, including \cite{wei2023surrounddepth, ding2024towards, kim2022self, guizilini2022full} and ours.
VFDepth achieves better results on DDAD compared to ours in multi-view consistency by processing features directly in 3D space. However, this approach and the other approaches underperform compared to ours in several examples, particularly when context in both images is different. This is especially evident along the boundaries of overlapping regions, where context from one image is incomplete, and in cases where object scales differ (see Fig.~\ref{fig:3d}). Moreover, our method has a considerably smaller memory footprint than VFDepth, as we operate on a two-dimensional cylindrical surface instead of 3D space (see Tab.~\ref{tab:memory}).
Similar is true for SurroundDepth, which relies on multi-head, purely learned attention with attention matrices eight times larger than ours. Yet, SurroundDepth underperforms compared to our non-learned geometry-based attention weighting, since purely learned attention does not guarantee feature aggregation from the correct tokens across images. CVCDepth faces a similar limitation, as multi-view consistency is only enforced implicitly through its loss functions. In contrast, our method, which follows CVCDepth with the addition of our proposed attention module, takes advantage of the known camera parameters to project all views into a shared cylindrical representation (see Fig.~\ref{fig:panorama}), explicitly ensuring multi-view consistency as illustrated by the attention weight maps in Fig.~\ref{fig:attn}. For more results, refer to the supp. material.

\subsection{Ablation Studies}

\begin{table}[t]
\centering
\setlength{\tabcolsep}{5pt} 
\renewcommand{\arraystretch}{0.95} 
\scriptsize
\resizebox{0.8\columnwidth}{!}{
\begin{tabular}{lcc}
\toprule
Method & Train [GB] & Inference [GB] \\
\midrule
FSM*                & 5.6 & \textbf{0.5} \\
VFDepth            & 11.0 & 3.3 \\
SurroundDepth      & 12.6 & 1.4 \\
CVCDepth           & \textbf{5.4} & 0.6 \\
\textbf{CylinderDepth (ours)} & 8.0 & 0.7 \\
\bottomrule
\end{tabular}%
}
\caption{Efficiency comparison of our method against state-of-the-art in terms of peak allocated memory during training and inference. FSM* denotes the implementation from \cite{kim2022self}.}
\label{tab:memory}
\end{table}

\begin{table}[t]
\centering
\setlength{\tabcolsep}{0.6pt} 
\renewcommand{\arraystretch}{0.9} 
\footnotesize  
\begin{tabular}{lcccccc}
\toprule
& \multicolumn{4}{c}{\textbf{Overall}} & \multicolumn{2}{c}{\textbf{Overlap}} \\
\cmidrule(lr){2-5} \cmidrule(lr){6-7}
Method & Abs Rel & Sq Rel & RMSE & $\delta < 1.25$ & Abs Rel & Depth Cons \\
\midrule
Ours (*)        & 0.212  & 3.741 & 13.21  & 70.0 & 0.214 &  \textbf{5.59}  \\
Ours (**)    & 0.208 & 3.500 & 12.90  & 70.2  &  0.211 & 6.72   \\
Ours (***)& 0.211 & 3.546  &  12.90  & 69.8 & 0.215 & 7.04  \\
Ours (****)    &  0.207  & 3.503 &  12.76 &   70.5 & 0.207  & 5.68   \\
Ours (*****)    &  0.204  & 3.422 &  12.73 &   70.8 & 0.205  & 6.05   \\
Ours (******)    &  \textbf{0.200}  & \textbf{3.272} &  \textbf{12.64} &   \textbf{71.2} & \textbf{0.202}  & 5.87   \\
\bottomrule
\end{tabular}
\caption{Ablation study on our method. (*) applying attention at all scales; (**) identity attention during training; (***) identity attention during inference with the full model; (****) our full model; (*****) MambaVision encoder without our proposed attention; (******) MambaVision encoder with our proposed attention. RMSE, Sq Rel and Depth Cons~are given in [m]. $\delta$ is given in [\%]. Abs Rel is unit-free. Results are reported for the entire images and for overlapping regions on the DDAD dataset.}
\label{tab:ablation}
\end{table}

To better assess our contribution and validate its effectiveness, we conduct ablation studies examining the impact of the proposed geometry-guided spatial attention during both training and inference, compare applying the attention only at a low scale versus at all scales, and analyze the role of the encoder design within our spatial attention.
\paragraph{Spatial Attention} To evaluate the influence of the proposed spatial attention mechanism, we keep the architecture unchanged and replace our attention weights (cf.~Eq.~\ref{eq:attention}) with an identity matrix, i.e., each token attends only to itself. We evaluate two settings: (i) identity-train, where the network is trained with identity attention, and (ii) identity-inference, where a model trained with our spatial attention is tested using identity attention. This study isolates the contribution of our spatial attention mechanism and demonstrates the benefit of cross-image feature sharing for multi-view consistency, particularly at inference (see Tab.~\ref{tab:ablation}).

\paragraph{Low-Scale Spatial Attention} We apply spatial attention only at the coarsest feature scale (cf.~Sec.~\ref{sec:method}), as cross-image attention behaves like a smoothing operator on the feature maps. By restricting attention to the lowest resolution, we enforce global multi-view consistency while preserving fine-scale structures in the higher-resolution features. In contrast, SurroundDepth applies attention at all scales by downsampling the high-resolution feature maps; for the ablation, we do the same. The predictions of this variant of our method exhibit reduced edge sharpness and appear over-smoothed, with slightly worse overall depth accuracy. Yet, the multi-view consistency does not improve significantly (see Tab.~\ref{tab:ablation} and supp. material).

\paragraph{Encoder Features} To further assess our contribution, we replaced the ResNet18 encoder with a recent state-of-the-art alternative, MambaVision-T~\cite{hatamizadeh2025mambavision}. We hypothesize that the depth inconsistency observed in the literature is not primarily attributable to the encoder architecture or its capacity. Instead, it arises from relying on a single shared encoder without any information exchange across images; consequently, the issue is largely agnostic to the specific encoder choice. This is reflected in Tab.~\ref{tab:ablation}. Specifically, the comparison of MambaVision with and without our attention mechanism shows the same behavior as for the used ResNet-18 encoder: the depth consistency improves only when features are explicitly shared via our attention mechanism.

%% file: sec/5_conclusions.tex
\section{Conclusion}
In this paper, we presented a method for self-supervised surround depth estimation, with a particular focus on enforcing multi-view consistency. Our approach projects pixels from all input images into a shared cylindrical representation, where attention is applied based on their distances on the cylinder. As shown by the results, this enables effective cross-image feature sharing, leading to improvements in multi-view consistency and overall depth accuracy. A limitation of the current design is that attention, due to its high computational cost, is applied only at the lowest feature resolution. While this enforces global consistency, the coarse scale aggregates large regions and restricts fine-grained detail, leading to suboptimal pixel-level consistency; we aim to address this issue in future work by adapting the distance computations. Moreover, we aim to model the rig's trajectory as a continuous function, instead of discrete time steps, to account for asynchronously taken images, as in nuScenes~\cite{caesar2020nuscenes}.